\documentclass[runningheads]{llncs}
\usepackage{graphicx}
\usepackage{comment}
\usepackage{array}
\usepackage{subfig}

\begin{document}
\title{{The effect of stemming and lemmatization on Portuguese fake news text classification}}
\titlerunning{The effect of stemming and lemmatization on fake news text classification}
%
\author{Lucca de Freitas Santos Calaigian\inst{1} \and
Murilo Varges da Silva\inst{1}}
\authorrunning{L.F.S. Calaigian and M.V. Silva}
%
\institute{Federal Institute of Education, Science and Technology of Sao Paulo, Birigui, Brazil \\
\email{freitas.s@aluno.ifsp.edu.br, murilo.varges@ifsp.edu.br}}
\maketitle              
\begin{abstract}
With the popularization of the internet, smartphones and social media, information is being spread quickly and easily way, which implies bigger traffic of information in the world, but there is a problem that is harming society with the dissemination of fake news. With a bigger flow of information, some people are trying to disseminate deceptive information and fake news. The automatic detection of fake news is a challenging task because to obtain a good result is necessary to deal with linguistics problems, especially when we are dealing with languages that not have been comprehensively studied yet, besides that, some techniques can help to reach a good result when we are dealing with text data, although, the motivation of detecting this deceptive information it is in the fact that the people need to know which information is true and trustful and which one is not. In this work, we present the effect the pre-processing methods such as lemmatization and stemming have on fake news classification, for that we designed some classifier models applying different pre-processing techniques. The results show that the pre-processing step is important to obtain betters results, the stemming and lemmatization techniques are interesting methods and need to be more studied to develop techniques focused on the Portuguese language so we can reach better results.

\end{abstract}
\section {Introduction}

The popularity of smartphones and social media is causing a great problem nowadays, the spreading of fake news. This kind of news can deceive thousands of people in a short time and harm not only the population but companies and society in general~\cite{silva2020towards}. According to \cite{burgoon1996interpersonal} fraud or deception is a type of information that is intentionally produced and shared with the goal of making a false impression or conclusion about a subject. Recently fake news is becoming one of the most dangerous deception sources, that is because this kind of information tries to mimic the content of the official press. However, is important to say that fake news is different from news that the source is not certain or that was not performed deep research on the subject, this kind of information is called misinformation~\cite{lazer2018science}. Consequently, this information can be misleading and even harmful, especially when it is disconnected from its origins and original contexts~\cite{rubin2014talip}. 

According to~\cite{gonzalez2003recuperaccao} the number of digital texts stored is growing faster, being forgotten because there is no one who can read and understand all these texts at once. To~\cite{chowdhury2003natural} the Natural Language Processing (NLP) is an area that explores the way that computers can be used to understand and manipulate the human language, being capable of doing useful tasks daily.

In Computer Science, NLP is not an easy task, because of the ambiguity of the natural language, this ambiguity makes NLP different from the processing of programming languages for example, which is defined as precisely avoiding ambiguity. NLP research can be focused on five levels of analysis (Phonetic or Phonological, Morphological, Syntactic, Semantic and Pragmatic) all levels have unique features and their own associated difficulties; however, each NLP application can be more focused on a subset of these levels~\cite{vieira2010processamento}.


Mining Text applications impose a hard restriction on the usual NLP Tools, as they involve large volumes of textual data, and they do not allow the integration of complex treatments (generating exponential algorithms, therefore, unapproachable). Besides that, the semantic models for the applications are rarely available, implying strong limitations on the identification of the semantic and pragmatic levels of the linguistic models \cite{rajman1998text}.

As believed by ~\cite{hauch2012linguistic} the automatic detection of deception information is highly important because of two facts: i. systems can be more objective than humans in verdicts, that’s because humans can be tendentious or be subject to bias and ii. who is judging can be overloaded and delayed or even commit a mistake on the verdict.

Applications based on NLP seek to use linguistic patterns that can help in the detection of fake news, as well as general deceptive information. However, there is some struggle with NLP research because it depends on a specific language, being necessary to restructure or develop new techniques to adapt the language of the study and even create specific techniques corpora to each language \cite{silva2020towards}.

In~\cite{rubin2015deception} the authors classified the deceptive information into three major types: i. Deception for humoristic purposes, using sarcasm and irony to develop parody and satire; ii. Deception content to fool the population and spread misinformation; and iii. Non-confirmed information that is publicly accepted

On~\cite{appling2015discriminative} a study is proposed to analyze the number of verbs, adjectives, and adverbs, the text complexity (average of the sentence size and average of the size of the words), the break (punctuations occurrence rate), incertitude (number of modal verbs and passive voices present on the text) and expressiveness (number of the adverbs and adjectives in relation to the number of nouns and verbs) to be possible to try recognize some patterns that can help on the automatic detection of deceptive information.

According to~\cite{appling2015discriminative} it is possible to encounter indications of falsifications, exaggeration, omission, and fraud in social media texts evaluating some points like lies, contradictions, distortions, superlatives, half-trues, phrase modification, irrelevant information, and misconceptions. Doing that is possible to detect automatically misleading information.

A recent survey \cite{oshikawa2018survey} describes the challenges involved in fake news detection using NLP and describes related tasks, in \cite{reis2019supervised} was presented a new set of features and measure the prediction performance of current approaches and features for automatic detection of fake news.

In \cite{cunha2020extended} the authors introduce three new steps into the pre-processing phase of text classification pipelines to improve effectiveness while reducing the associated costs. Text pre-processing and normalizations are techniques that seek to decrease e standardize textual elements to facilitate information retrieval~\cite{aw2006phrase}. 

Among the existent techniques in the literature, this paper focuses on Lemmatization and Stemming, both techniques aim to reduce the word to its root to decrease the existent dictionary size.

Stemming has been the most widely applied morphological technique for information retrieval, stemming reduces the total number of index entries, however, stemming causes query expansion by bringing word variants and derivations~\cite{korenius2004stemming}. 

Lemmatization is like stemming having the same benefits, however, when this technique is applied, the lemma of the word is extracted, this lemma has the purpose of being a real world that exists in the language in study \cite{korenius2004stemming,aw2006phrase}. 

A lemmatization problem studied by~\cite{matthews2014concise} happens when the lemmatization technique finds a compound word. Many times, the algorithm can’t find a difference between the compound word or its component of the word unattended, this problem is very important when the goal of the research is text information retrieval.

In the study~\cite{korenius2004stemming} the authors suggest that the lemmatization algorithm tends to have a better result, that’s because the stemming techniques have a bigger recall and smaller precision than the lemmatization, although recall has an important weight on efficiency statistics, lemmatization show yourself superior because its metrics are balanced.
 
In~\cite{shu2017fake} the authors proposed some methods for detecting fake news, the first method is knowledge-based, which means that the data used to train the artificial intelligence model is labeled on true and fake news, other method used is style-based, this method tries to detect a pattern of writing that seeks to fool the people. On the study of \cite{fakebr:18} was found that the true news has more words than the fake news, being that a pattern that can be analyzed in detail.
 
On the research of~\cite{silva:20} the SVM, Logistic Regression and Random Forest algorithms are shown as the better ones for fake news classification, however, they say that there was no superior method, but the ones that got better results than the others.

The goal of this paper is present a study of the effect of lemmatization and stemming techniques on the classification of Fake News. To reach this objective were created some classification models that went analyzed and explored to appoint which technique obtained the best result and explore the reasons for it. 

Besides this introductory section, this paper is organized as follows. Section \ref{sec:proposed}  presents the lemmatization and stemming methods used in this work. Section \ref{sec:experiments} presents the experiments and the discussions of the results obtained by applying the methods of text normalization previously mentioned. Section \ref{sec:conclusion} presents the conclusion of the work.

\section{Proposed Method}
\label{sec:proposed}

\begin{figure}[htb]
\includegraphics[scale=.65]{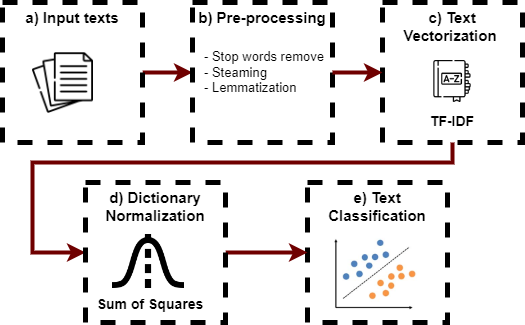}
\centering
\caption{Illustration of the main steps for the proposed method.} \label{fig:fakenewsflowchart}
\end{figure}

In order to classify texts as fake or true news class was proposed a method that is composed of five main steps, as follows: 

\renewcommand{\labelenumi}{\alph{enumi})}
\begin{enumerate}
    \item Reading input news texts (fake or true);
    \item Applying preprocessing methods (stop-words removal, steaming and lemmatization normalization);
    \item Text vectorization using TF-IDF (Term Frequency-Inverse Document Frequency);
    \item Dictionary normalization using the Sum of Squares;
    \item Text classification using classification algorithms.
\end{enumerate}
 This proposed method was tested utilizing a database of news in Portuguese labeled as fake or true, Figure \ref{fig:fakenewsflowchart} shows the main steps of the proposed method.

The preprocessing step (b) in Figure \ref{fig:fakenewsflowchart} consists in removing the stop-words and applying the text normalization techniques (Steaming and Lemmatization). It is important to say that we applied these techniques separately, we will discuss this in the experiments section. The stop-words were removed by the authors of the database \cite{fakebr:18}, for the stemming and lemmatization normalization techniques, it was used two Python libraries focused on NLP problems, NLTK~\cite{bird2006nltk} and SPACY~\cite{srinivasa2018natural}, they were chosen because they have methods for the Portuguese Language. 

Text vectorization is the process of converting text into numerical representation ~\cite{singh2019vectorization}. The technique for text vectorization (step (c) in Figure \ref{fig:fakenewsflowchart}) that we used in our proposed method is the TF-IDF (Term Frequency-Inverse Document Frequency), this technique consists of a statistical measure that can indicate the magnitude of a word on a document in relation to a set of documents or a linguistic corpus, as shown in Equation \ref{eq:tfidf}:

\begin{equation}
\label{eq:tfidf}
    TFIDF = TF \times IDF
\end{equation}

$TF$ can be calculated as:

\begin{equation}
    TF = \frac{N(t)}{T(t)}
\end{equation}

Where $N(t)$ is the number of times that a term $t$ appears in a document and $T(t)$ is the total number of terms in the document. Now we can compute the $IDF$ as follows:

\begin{equation}
    IDF = \frac{\log T(n)}{N(t)}
\end{equation}

$T(n)$ is the total number of documents and $N(t)$ is the number of documents with the term $t$ in it. 

The TF-IDF word value increases proportionally as the number of occurrences of it in a document increases, although, this value is balanced by the frequency of this word on the corpus, this helps to deal with the stop-words or the words that are more common than others \cite{aizawa2003information}. 

The dictionary normalization step (d) in Figure \ref{fig:fakenewsflowchart} consists in applying normalization techniques to facilitate the analysis and application of methods. To normalize the dictionary, it was used the sum of squares method that is present in the Scikit-Learn library \cite{pedregosa2011scikit}.

For the classification, step (e) in Figure \ref{fig:fakenewsflowchart}, used three common classifiers: SVM (Support Vector Machine), KNN (K-Nearest Neighbors) and DT (Decision Tree). The SVM classifier was used with Linear Kernel, the KNN one was set for three neighbors on the neighborhood and for the Decision Tree it was chosen a max of three leaves.

\section{Experiments and Results}
\label{sec:experiments}

For applying the proposed method, we use the \footnote{https://research.google.com/colaboratory/}{Google Colab} platform, which Google makes available infrastructure with Python and Jupyter Notebooks for free, this platform is widely used for studies, research and development in the data mining and machine learning fields.

The used database is Fake.Br Corpus, which is compounded by news written in Portuguese, is mostly political news \cite{fakebr:18}. However, it contains some other news, being in a total of 7200 news, this database is balanced, i.e., half of the news (3600) are labeled fake news and the other half (3600) are labeled true news. The Fake.Br Corpus database was passed through a pre-processing method by the authors of it, the authors removed stop-words, such as accents and diacritics. The headlines, titles, videos and images that news can contain were not considered for detection in this paper, we used only the text that tells the news.

\begin{figure}[htb]
\includegraphics[scale=.4]{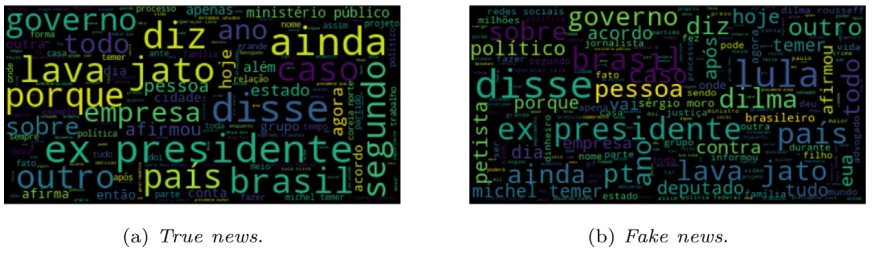}
\centering
\caption{Word clouds representing the relative frequency of the tokens, available in \cite{fakebr:18}.} 
\label{fig:wordcloud}
\end{figure}

\begin{figure}[htb]
\includegraphics[scale=.4]{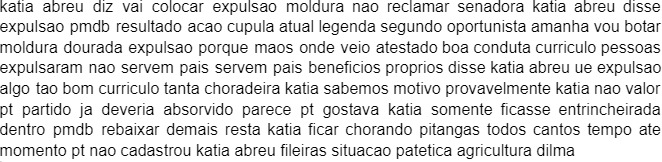}
\centering
\caption{Example of a new included on the dataset (Without stop-words).} 
\label{fig:textexample}
\end{figure}

Figure \ref{fig:wordcloud} shows the most frequent words present on the database and Figure \ref{fig:textexample} displays an example of one news that is included in the dataset.

For the train and test step the database was split, 30\% of all data were used to perform the tests of the models and 70\% were used to train the classifier algorithms. Table \ref{tab:experiments_setup} shows the setup and steps that were used to perform the experiments. 

\begin{center}

\begin{table}[htb]
\centering
\caption{Experiments setup and steps used to perform the experiments.}
\label{tab:experiments_setup}
\begin{tabular}{|m{5cm}|m{5cm}|}
\hline
\textbf{Experiment name} & \textbf{Experiment setup description}
\\ \hline
Stop-Words Removal & It was used database collected without applying any of the normalization methods. 
\\ \hline
Stop-Words Removal + Dictionary of 500 words       & The dictionary created  was limited by 500 words.                                 \\ \hline
{Stop-Words Removal + Dictionary of 5000 words}      & The dictionary created was limited by 5000 words                                  \\ \hline
{Lemmatization Technique}                            & Applied the Lemmatization technique                                               \\ \hline
{Lemmatization Technique + Dictionary of 500 words}  & Dictionary created limited by 500 words for the lemmatization technique           \\ \hline
{Lemmatization Technique + Dictionary of 5000 words} & Dictionary created limited by 5000 words for the lemmatization technique          \\ \hline
{Stemming Technique}                                 & Applied the Stemming technique                                                    \\ \hline
{Stemming Technique + Dictionary of 5000 words}      & Dictionary created limited by 500 words for the stemming technique                \\ \hline
{Stemming Technique + Dictionary of 500 words}       & Dictionary created limited by 5000 words for the stemming technique                \\ \hline
\end{tabular}
\end{table}
\end{center}

To perform the tests, we applied both methods of text normalization (Lemmatization and Stemming) to the data, after that we build some different dictionaries, but used the same methodology. The first list of dictionaries created was compound of all words that the vectorization algorithm judged relevant, then the second list was created by dictionaries that were limited by 500 words, this value was chosen to reduce the dimension of the model and analyze the impact of these limitations on data classification. The third list of dictionaries created was thought similarly to the second list, but now the dictionaries created were limited by 5000 words, again this value was chosen to analyze the impact of a medium-size dictionary on the data classification. The idea behind switching the values of the words that compound the dictionary is that we can reduce the complexity and the dimension of the models and get good results, or at least the same results as other complex models. Table \ref{tab:dic_size} presents the number of words presents on the first list of dictionaries created by each method in the study.

\begin{table}[htb]
\caption{Size of the built dictionaries applying each experiment setup.}
\label{tab:dic_size}
\centering
\scalebox{1}{
\begin{tabular}{| c | c |}
\hline
\textbf{Method} & \textbf{Dictionary size}\\ \hline
{Stop-Word Removal}                      & 68562 words                     \\ \hline
{Lemmatizing}                            & 57086 words                     \\ \hline
{Stemming}                               & 31485 words                     \\ \hline
\end{tabular}%
}
\end{table}

For each list of dictionaries, we applied the classifier algorithms quoted previously. Table \ref{tab:results} shows de results of the experiments.

\begin{table}[htb]
\centering
\caption{Accuracy rates (\%) for fake news classification using FAKE.Br Corpus Dataset, using classifiers: SVM, KNN and Decision Tree (DT).}
\label{tab:results}

\scalebox{1.0}{
\begin{tabular}{|m{8cm}|c|c|c|}

\hline
\textbf{Experiment name}  & \textbf{SVM}   & \textbf{KNN}   & \textbf{DT} \\ \hline
Stop-Words Removal                                 & 96.11 & 71.06 & 85.79         \\ \hline
Stop-Words Removal + Dictionary of 500 words       & 93.75 & \textbf{74.95} & 85.79         \\ \hline
Stop-Words Removal + Dictionary of 5000 words      & \textbf{96}.20 & 72.31 & 85.79         \\ \hline
Lemmatization Technique                            & 95.25 & 69.72 & 83.80         \\ \hline
Lemmatization Technique + Dictionary of 500 words  & 94.16 & 73.19 & 83.80         \\ \hline
Lemmatization Technique + Dictionary of 5000 words & 95.87 & 70.83 & 83.80         \\ \hline
Stemming Technique                                 & 95.69 & 70.83 & 86.06         \\ \hline
Stemming Technique + Dictionary of 500 words       & 93.70 & 71.62 & 86.06         \\ \hline
Stemming Technique + Dictionary of 5000 words      & 96.11 & 70.79 & 86.06         \\ \hline
\end{tabular}%
}
\end{table}

\begin{figure}[htb]
  \centering
  \subfloat[Lemmatizing.]{\includegraphics[width=0.5\textwidth]{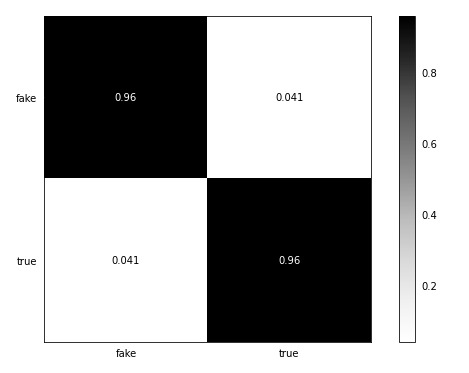}\label{fig:cm_lemma}}
  \hfill
  \subfloat[Stemming.]{\includegraphics[width=0.45\textwidth]{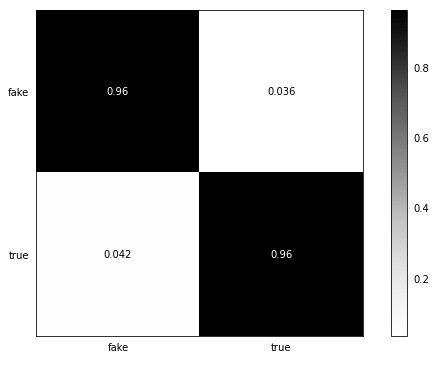}\label{fig:cm_steam}}
  \caption{Confusion Matrix of the best results 5000-words. (a) Lemmatizing technique and (b) Stemming technique. }
\label{fig:cm_lemma_steam}
\end{figure}

As we can see, better results were obtained using the SVM classifiers. Figure \ref{fig:cm_lemma_steam} shows the confusion matrices that represent the best result for the stemming and lemmatization techniques. 

Analyzing Table \ref{tab:results} we can see that for the lemmatization, the better result was obtained by using a dictionary with 5000 words, Figure \ref{fig:cm_lemma} shows the confusion matrix of it. 

In the same way, we can see that the stemming technique got a better result when using the 5000-word dictionary again. Figure \ref{fig:cm_steam} exposes the confusion matrix for this method.

\begin{figure}[htb]
  \centering
  \subfloat[Only Stop-words.]{\includegraphics[width=0.5\textwidth]{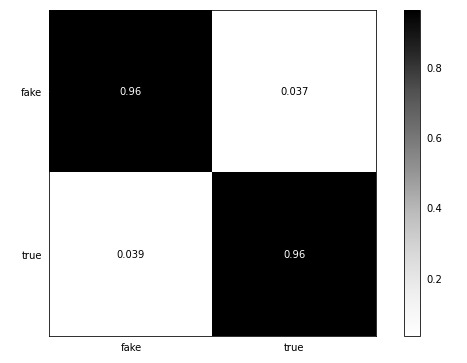}\label{fig:cm_stopwords}}
  \hfill
  \subfloat[Lemmatization.]{\includegraphics[width=0.48\textwidth]{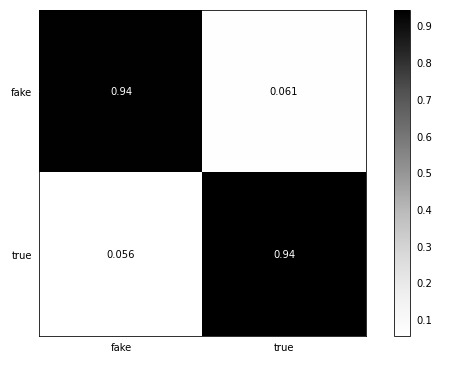}\label{fig:lemma_500}}
  \caption{Confusion matrix obtained by the SVM model. (a) Only stop-word (5000-word) and (b) Lemmatization technique (500-word). }
\end{figure}

The best method overall got 96.20\% accuracy and was obtained by using only the Stop-Words removal and a dictionary of 5000 words. As we can notice, all three methods (Stop-Words Removal, Lemmatizing and Stemming) had the best results when using a 5000-word dictionary. This is an interesting result when we think that this dictionary is at least 6 times smaller than the one built without limitations. Figure \ref{fig:cm_stopwords} shows the confusion matrix for this method. 

To end the discussions section,  we can notice that for the 500-word dictionary, the best result was obtained by the lemmatization technique, that is an exciting result because for the other dictionaries, we can notice that only the stop-word removal pre-process was getting better results. This is an exciting result because we can see that the lemmatization technique can have good results, especially when we compared our results to other state-of-the-art works that say that the lemmatization technique produced the best overall results. Figure \ref{fig:lemma_500} shows the confusion matrix built by this method.

\section{Conclusion}
\label{sec:conclusion}
The spread of fake news is an ordinary problem these days, and fighting against it is extremely important for society, that's because people need to take decisions and attitudes based on true facts and situations. The development of methods that can automatically and safely detect fake news shows very important since individuals can't or at least don't try to check the veracity of the information. 

This paper presents a new study of the treatment of text data using stemming and lemmatization methods and techniques to improve the results of the automatic detection of fake news in the Portuguese language. Accordingly, with the obtained results, we can observe that the dictionaries that have an extensive group of words provides better results than the dictionaries that have lesser words. The stemming and lemmatization methods used showed themselves promising for the solution of the problem, however, we can notice that the results do not differ from the results obtained without applying any normalization method, it is worth mentioning that the stemming method reduced halved the words the size of the dictionary and even got a good result, this is a positive accomplish for this experiment. 

Most text normalization techniques (Stemming and Lemmatization) were developed to be used in the English language, therefore, they do not perform well when applied to texts written in Portuguese. The text normalization area needs more research directed to the Portuguese language.

Finally, the obtained results are satisfactory, especially when compared to related works, in which we can notice some improvement in some methods and results.
%
%
%
\bibliographystyle{splncs04}
\bibliography{ciarpbib}

\end{document}